\documentclass[runningheads]{llncs}
\usepackage[T1]{fontenc}
\usepackage{graphicx,verbatim}
\usepackage{hyperref}
\usepackage{color}

\usepackage{ifthen}
\usepackage{amsmath, amsfonts, amssymb}
\usepackage{float}
\usepackage{booktabs}
\usepackage{subfig}
\usepackage{enumitem}
\usepackage{bbm}
\usepackage{gensymb}
\usepackage{mathtools}
\usepackage{multirow}
\usepackage{multicol}
\usepackage{xspace}
\usepackage[square,numbers,sort&compress,sectionbib]{natbib}

\makeatletter
\DeclareRobustCommand\onedot{\futurelet\@let@token\@onedot}
\def\@onedot{\ifx\@let@token.\else.\null\fi\xspace}
\def\eg{\emph{e.g}\onedot}

\newlist{inlist}{enumerate*}{1}
\setlist[inlist]{label=(\arabic*)}
\newlist{inlistalpha}{enumerate*}{1}
\setlist[inlistalpha]{label=(\alph*)}

\newcommand{\para}[1]{\noindent\textbf{#1}.~~}
\begin{document}
\title{FluoroSAM: A Language-promptable Foundation Model for Flexible X-ray Image Segmentation}
\titlerunning{FluoroSAM}

\author{Benjamin~D.~Killeen\inst{\ast 1} \and
Liam~J. Wang\inst{1}\and
Blanca I\~n\'igo\inst{1}\and
Han Zhang\inst{1}\and
Mehran Armand\inst{2}\and
Russell H.~Taylor\inst{1}\and
Greg Osgood\inst{3}\and
Mathias Unberath\inst{1}
}
\authorrunning{B.D. Killeen et al.}
%
\institute{Johns Hopkins University, Baltimore, MD 21218, USA\\
\email{\{killeen, wwang136, binigo1, hzhang206, rht, unberath\}@jhu.edu}\\
\and University of Arkansas, Fayetteville, AR 72701, USA\\
\email{marmand@uark.edu}\\
\and Johns Hopkins Hospital, Baltimore, MD 21287, USA\\
\email{\{gosgood2\}@jhmi.edu}}


\maketitle              
\begin{abstract}
  Language promptable X-ray image segmentation would enable greater flexibility for human-in-the-loop workflows in diagnostic and interventional precision medicine. Prior efforts have contributed task-specific models capable of solving problems within a narrow scope, but expanding to broader use requires additional data, annotations, and training time. Recently, language-aligned foundation models (LFMs) -- machine learning models trained on large amounts of highly variable image and text data thus enabling broad applicability -- have emerged as promising tools for automated image analysis. Existing foundation models for medical image analysis focus on scenarios and modalities where large, richly annotated datasets are available. However, the X-ray imaging modality features highly variable image appearance and applications, from diagnostic chest X-rays to interventional fluoroscopy, with varying availability of data. To pave the way toward an LFM for comprehensive and language-aligned analysis of arbitrary medical X-ray images, we introduce FluoroSAM, a language-promptable variant of the Segment-Anything Model, trained from scratch on 3M synthetic X-ray images from a wide variety of human anatomies, imaging geometries, and viewing angles. These include pseudo-ground truth masks for 128 organ types and 464 tools with associated text descriptions. FluoroSAM is capable of segmenting myriad anatomical structures and tools based on natural language prompts, thanks to the novel incorporation of vector quantization (VQ) of text embeddings in the training process. We demonstrate FluoroSAM's performance quantitatively on real X-ray images and showcase on several applications how FluoroSAM is a key enabler for rich human-machine interaction in the X-ray image acquisition and analysis context. \textit{Code is available at \url{https://github.com/arcadelab/fluorosam}.}

\end{abstract}

\section{Introduction}

X-ray imaging is a workhorse imaging modality for diagnostic and interventional healthcare. There is enormous opportunity for language-promptable, automated segmentation of X-ray images to enable human-in-the-loop workflows in precision medicine.\cite{calli2021deep, zhao2021artificial, killeen2023silico, killeen2023pelphix, unberath2018deepdrr, gao2023synthetic, silva2018automatic, candemir2014lung} Prior efforts have contributed machine learning (ML) techniques that perform well within a narrow scope of X-ray imaging, but their fixed design and limited training data limit the potential across the domain. Extending these techniques to support additional classes or more complex queries requires additional data and personnel effort for refitting and retraining ML models. Recently, foundation models (FMs) -- particularly language-aligned FMs (LFMs) -- have emerged as a promising direction for building more flexible models~\cite{ma2024segment, brown2020language, Liu2023CVPR, chen2024chexagent, kirillov2023segment}. FMs are characterized by scalable training strategies, often accomplished through self-supervision, that enable learning from large, diverse data. LFMs, which incorporating text into the training process alongside images, allow for specification of tasks and classes using natural language. Our goal is to develop a language-promptable FM for X-ray image segmentation, enabling potential downstream applications ranging from interactive diagnostic systems~\cite{jiang2024comt,chen2024chexagent} to intelligent human-machine interfaces in image-guided interventions~\cite{killeen2024take,killeen2025intelligent}.

\begin{figure}[t]
  \centering
  \includegraphics[width=0.99\linewidth]{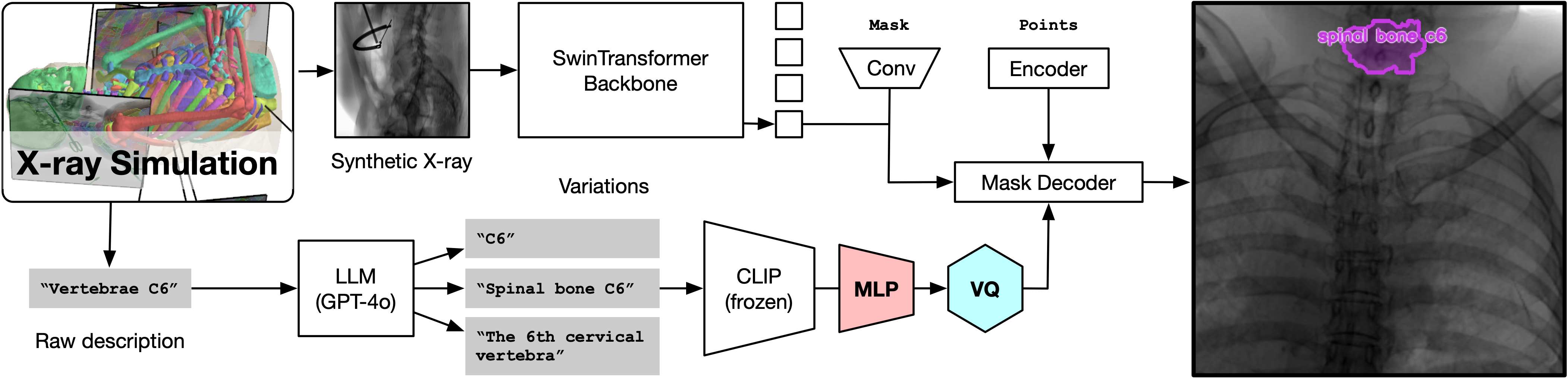}
  \caption{\textbf{FluoroSAM} is trained fully \emph{in silico} with mask and text prompt pairs. It features a VQ layer that enables training language-aligned training on X-ray images.}
  \label{fig:model}
\end{figure}
Promptable segmentation models have become increasingly prominent in recent years, as they are adaptable to both automated and human-in-the-loop workflows. Rather than predict segmentation masks for a fixed number of classes, the Segment-Anything Model (SAM) \cite{kirillov2023segment} predicts a semantically meaningful mask for any object in an image based on a prompt, which can be a mask, bounding box, or point. Trained on a large scale dataset of over 1B masks, the original SAM is a powerful tool for automated and interactive image segmentation, and it has been successfully fine-tuned on a variety of medical imaging modalities~\cite{ma2024segment}. Its successor, SAM 2, further lends itself to 3D medical image segmentation by incorporating a memory bank, increasing the efficiency and consistency of segmentation over multiple frames~\cite{ravi2024sam}. However, to the extent that SAM and SAM 2 support text prompts, they rely on a previously trained LFM, CLIP~\cite{radford2021learning}, to convert image patches into language-aligned prompts. It is not immediately clear how to transfer this approach outside of natural images, on which CLIP was trained. Whereas natural images -- and many medical imaging modalities -- feature objects with clear boundaries and a nested structure, X-ray images are transmissive by nature, with many overlapping masks belonging to wholly different objects. Even if a CLIP-like model were available for the X-ray imaging modality, a single image patch would contain visual features from multiple objects, and it would be unclear how to distinguish among them. Further, the data available for training LFMs for X-ray imaging has so far been limited to diagnostic chest X-ray, where the imaging geometry is relatively consistent and detailed radiology reports can be retrospectively sourced for training data~\cite{chen2024chexagent,irvin2019chexpert,candemir2014lung}. By contrast, interventional X-ray features a wide variety of imaging geometries, anatomies, and objects, with little data available for training.

\begin{figure}
  \centering
  \includegraphics[width=0.8\linewidth]{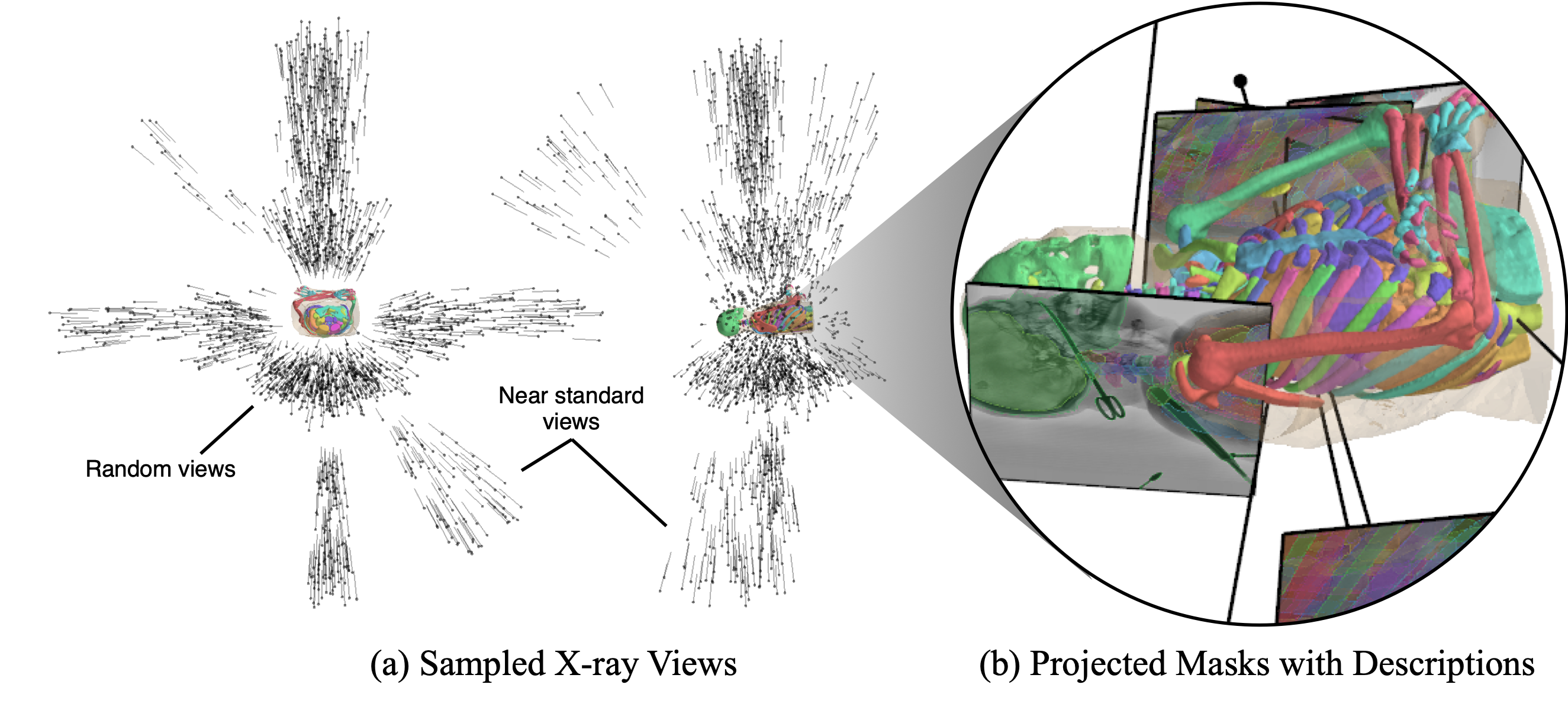}
  \caption{\textbf{Simulation of the FluroSeg dataset.} (a) Virtual C-arm views, shown here as principle rays, are sampled from a range of both random and near-standard views, which are determined automatically. (b) The environment outputs masks and descriptions for organs and tools.}
  \label{fig:views}
\end{figure}

Recent advances in simulation~\cite{unberath2018deepdrr}, automated CT image segmentation~\cite{wasserthal2023totalsegmentator}, and sim-to-real transfer~\cite{gao2023synthetic, killeen2023pelphix} introduce the possibility of training LFMs for X-ray image segmentation with paired masks and text descriptions.
To enable flexible natural language prompting, we use a vector quantization (VQ) module on top of a frozen CLIP encoder to provide a more consistent signal to the mask decoder.
In this context, we make two major contributions:
\begin{enumerate}
  \item The FluoroSeg dataset, a large-scale dataset of 3M synthetic X-ray images, with mask and text pairs for organs and tools. The dataset is generated from a wide variety of human anatomies, imaging geometries, and viewing angles, and it is designed to support training of a language-promptable FM for X-ray image segmentation.
  \item FluoroSAM, a language-promptable variant of SAM, trained from scratch on FluoroSeg. FluoroSAM features a novel text encoder that enables generalizable language alignment during training. In this paper, we focus our evaluation on real fluoroscopic images and chest X-rays, for which ground truth segmentations are available.
\end{enumerate}

\section{Methods}
\para{FluoroSeg Dataset}
FluoroSeg is a large-scale dataset of 3 million simulated X-ray images of the full body, for both diagnostic and interventional exams. Building on prior work~\cite{unberath2018deepdrr,unberath2019enabling, killeen2023pelphix, gao2023synthetic}, the simulation environment takes as input a patient model derived from a real CT scan, along with associated segmentations and descriptions. Computer modeled surgical tools are positioned relative to the patient anatomy. Where FluoroSeg stands out in this space is in the scalability of image and ground truth simulation as well as the number of radiological exams which are replicated in a given environment. By combining volumetric and mesh-based rendering methods for CT, tool, and ground truth projection, the FluoroSeg simulation environment is able to generate numerous images with myriad ground truth segmentations and descriptions in an efficient, scalable manner. To replicate the variability of real-world X-ray imaging, the simulation environment samples from imaging geometries and viewing angles that are both random and near-standard, as determined automatically based on organ meshes.

The simulation pipeline is as follows. We source 1621 high resolution ($0.96 \times 0.96 \times 0.5$\,mm) CT scans from the New Mexico Decedent Image Database~\cite{edgar2020new}, including scans of the head \& neck region, the torso, and the lower extremity. Each scan is segmented into 128 organ classes using TotalSegmentator~\cite{wasserthal2023totalsegmentator}, from which we obtain surface meshes. Depending on the scan type, the simulation environment selects a range of standard views from which to render synthetic X-ray images, including the chest, abdominal, shoulder, clavicle, humerus, elbow, forearm, hand, pelvis, femur, sacroiliac joint, knee, tibia/fibula, ankle, foot, skull, and spine series as defined by \cite{murphy2020general}. Each view is approximated with minor random variations based on the meshes present, so as to capture a variety of imaging geometries for each exam, suited to both diagnostic and interventional applications. In addition, fully random C-arm views are sampled by selecting a primary organ to focus on and then randomly selecting a viewing angle to within 60\degree of the anterior direction. This strategy is reflected in Fig.~\ref{fig:views}, in which the sampled source positions and principle ray directions can be seen clustered around near-standard views as well as distributed randomlyk.
For each image, we select a random subset out of 464 tools to include in the image. These include 111 models from \cite{luijten20233d}, 296 from GrabCAD, and 57 modeled internally. Each tool is associated with a comprehensive text description, \eg ``cannulated 110mm screw.'' They are placed along the field of view with random location and orientation. The number of images sampled per CT varies based on the number of standard views which are possible to determine for the given anatomy.
Images are rendered at a resolution of $512^2$ pixels alongside masks and text descriptions. Using an NVIDIA A6000 GPU, the simulation environment is able to generate $6.5 \pm 15.7$ images per second, depending on the exam, with a total of 2.95M images generated in about 6 GPU days.
The final images are split into training and validation sets based on a 90/10\% split of the base CT scans.


\begin{figure}[t]
  \centering
  \includegraphics[width=\linewidth]{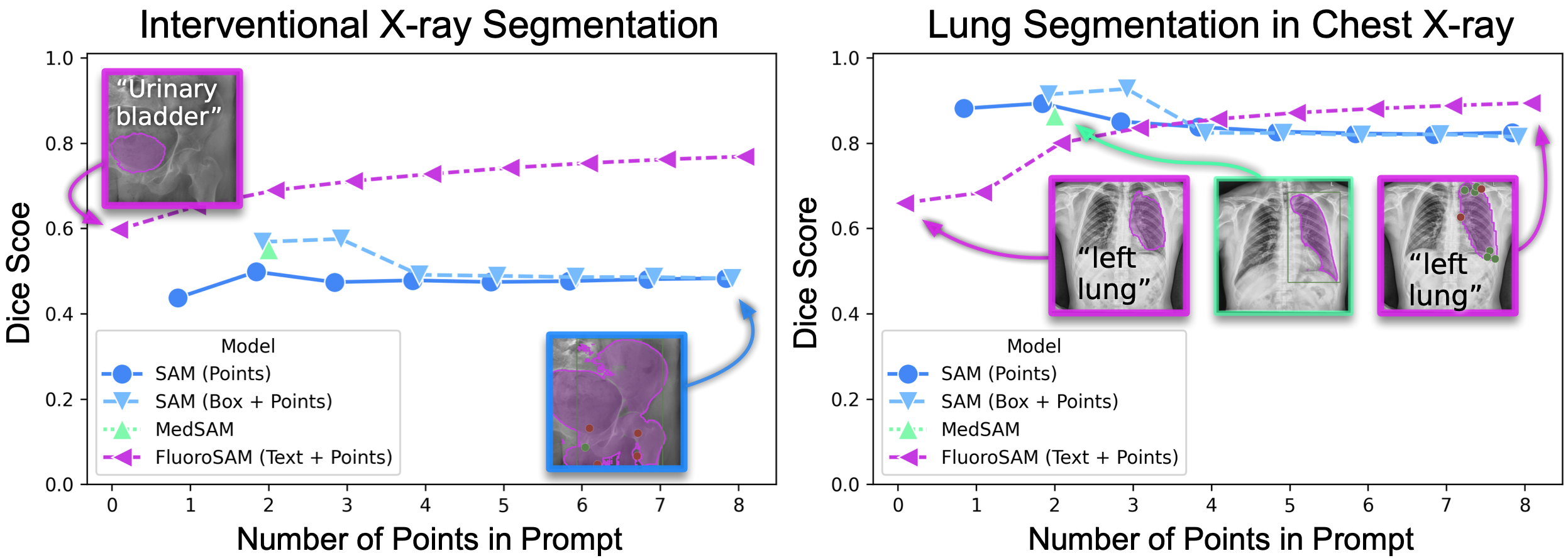}
  \caption{\textbf{Quantitative results}. (a) On interventional X-ray images, FluoroSAM outperforms its peers even with text-only prompting. (b) On CXRs, FluoroSAM adapts to hand-annotated lung segmentations despite being trained on synthetic data. MedSAM~\cite{ma2024segment} includes this task in its training data.}
  \label{fig:plots}
\end{figure}

\para{FluoroSAM: An LFM for X-ray Image Segmentation}
FluoroSAM is a language-promptable variant of SAM~\cite{kirillov2023segment} with support for text, point, and mask prompts, as shown in Fig.~\ref{fig:model}. When using only a single prompt, SAM predicts three segmentation maps, often corresponding to the whole, part, or sub-part of an object, backpropagating only through the branch with the lowest loss. X-ray images, on the other hand, often contain overlapping projections of many anatomical and non-anatomical objects. MedSAM~\cite{ma2024segment} mitigates this ambiguity for other medical imaging domains by only allowing for bounding box prompts. For X-ray imaging, even if MedSAM were trained from scratch using the FluoroSeg dataset, this approach is undesirable because it
\begin{inlistalpha}
\item still features significant ambiguity,
\item makes automatic or even non-expert prompting impractical, and
\item does not lend greater flexibility for human-in-the-loop systems.
\end{inlistalpha}
Language prompts, on the other hand, enable greater flexibility for both automated and human-in-the-loop systems, while also reducing ambiguity in prompts.

With access to the text descriptions in the FluoroSeg dataset, it is possible to train FluoroSAM from scratch with text prompting, using a novel text encoder. The text encoder consists of a frozen CLIP encoder~\cite{radford2021learning}, followed by a multi-layer perceptron (MLP) with 2 hidden layers and a VQ bottleneck~\cite{oord2017neural}, which outputs the prompt token. While VQ theoretically limits the generalizability of our framework to new segmentation classes, it expands the generalizability to new language prompts by reducing the variability from disparate descriptions for the same object. During training, we use \texttt{gpt-4o}~\cite{brown2020language} to perform text augmentation~\cite{fan2023improving} on the original descriptions, generating up to 30 comprehensive and non-comprehensive prompts for each mask. In addition, we procedurally combine masks from related organs in 38 groups, such as the ``left ribs'' or ``cervical vertebrae,'' which are separate classes in the base TotalSegmentator, and likewise augment their descriptions. Finally, with small probability, we sample text prompts which are not present in the image, so that FluoroSAM ignores bad prompts. Because text prompts themselves may be ambiguous, we leverage the multi-output capabilities of SAM to predict multiple masks for each prompt, and we select the mask with the lowest loss during training. During inference, the IOU prediction head enables selection of the best mask for each prompt.

The image encoder consists of a SwinTransformer backbone, which is better suited to the simulated image size than the original ViT backbone, pre-trained on ImageNet-22k, with additional pre-training for instance segmentation on a reduced set of FluoroSeg classes. We use the Swin-S variant for ablation studies and Swin-L for the final model, with an input image size of $448^2$. Following \cite{gao2023synthetic,killeen2023pelphix}, we apply strong domain randomization of the image during training to facilitate sim-to-real transfer, including coarse dropout, inversion, blurring, Gaussian contrast adjustment~\cite{killeen2023corridors}, random windowing, and CLAHE histogram equalization. To accommodate the wide variety of intensity values in X-ray images, we used a 3-channel input, where the second and third channel window and level are determined by a K-Means clustering of the pixel values. We use Dice and focal loss for masks, re-weighting the contributions from text-only prompting to equal that of all point prompts. We train FluoroSAM for 10 epochs with a base learning rate of $8 \times 10^{-4}$ after a linear warm-up of 20k iterations from $8 \times 10^{-6}$, reduced by a factor of 10 at step 200k and 400k. We use a batch size of 16 across 2 NVIDIA H100 GPUs, with a total training time of 6 days, and up to 8 point prompts per image.
\section{Evaluation}

\begin{table}[t]
  \centering
  \caption{\textbf{Real X-ray results.} (*) indicates that a few ($< 10$) false negatives were removed to compute HDD. HDD is in mm for cadaver, pixels for CXRs.}
  \footnotesize
  \begin{tabular}{llrrrrrrrrrr}
    \toprule
    Dataset & Model & \multicolumn{3}{c}{Text Only} & \multicolumn{3}{c}{2 Points or Box} & \multicolumn{3}{c}{8 Points}\\
    \cmidrule(lr){3-5} \cmidrule(lr){6-8} \cmidrule(lr){9-11}
    && IoU$\uparrow$ & Dice$\uparrow$ & HDD$\downarrow$ & IoU$\uparrow$ & Dice$\uparrow$ & HDD$\downarrow$ & IoU$\uparrow$ & Dice$\uparrow$ & HDD$\downarrow$ \\
    \midrule
    Cadaver & SAM &  & --- &  & 0.36 & 0.50 & 139.1* & 0.35 & 0.48 & 152.6* \\
    & SAM (Box) &  & --- &  & \underline{0.42} & \underline{0.57} & 86.6* & \underline{0.35} & \underline{0.48} & \underline{143.9}* \\
    & MedSAM &  & --- &  & 0.41 & 0.55 & \textbf{76.0} &  & --- &  \\
    & FluoroSAM & \textbf{0.47} & \textbf{0.60} & \textbf{102.8}* & \textbf{0.56} & \textbf{0.69} & \underline{81.0} & \textbf{0.64} & \textbf{0.77} & \textbf{60.8} \\
    \cmidrule(lr){1-1} CXR & SAM &  & --- &  & \underline{0.83} & \underline{0.89} & 153.9 & \underline{0.73} & \underline{0.82} & 406.8 \\
    & SAM (Box) &  & --- &  & \textbf{0.85} & \textbf{0.91} & 87.6 & 0.73 & 0.81 & \underline{303.9} \\
    & MedSAM &  & --- &  & 0.34 & 0.48 & \underline{81.9} & & --- & \\
    & FluoroSAM & \textbf{0.50} & \textbf{0.66} & \textbf{90.1} & 0.67 & 0.80 & \textbf{53.9} & \textbf{0.81} & \textbf{0.89} & \textbf{31.9} \\
    \bottomrule
  \end{tabular}
  \label{tab:results}
\end{table}

\begin{figure}[t]
  \centering
  \includegraphics[width=0.9\linewidth]{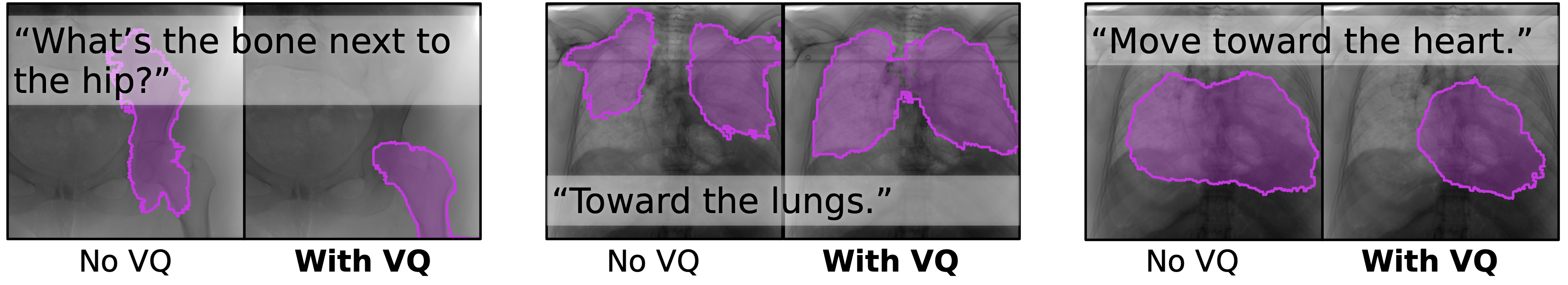}
  \caption{In a limited user study, we observe qualitative results consistent with the hypothesis that VQ improves segmentation robustness to variable text prompts. For example, FluoroSAM with VQ was able to correctly segment the femur, answering the question ``What's the bone next to the hip?''}
  \label{fig:ablation}
\end{figure}
\para{Synthetic X-ray images} Evaluated on the FluoroSeg dataset, FluoroSAM achieves a Dice score of 0.59 based solely on text prompting, increasing to 0.70 with 2 points and 0.79 with 8 points. This is a promising result given the challenge of full body X-ray segmentation based on text prompts.

\para{Real Interventional X-ray images}
To evaluate FluoroSAM's performance on real interventional X-ray iamges, we collect a dataset of registered X-ray images with a full torso specimen using a Brainlab Loop-X device.
The specimen consisted of a torso section from mid-femur to T2, excluding the arms, obtained from a 60-year-old female donor with a living height of 157\,cm and weight of 50\,kg. Before the study, the specimen was thawed at 4\degree C for six days. All fluoroscopic images were acquired with navigation relative to a fixed patient array. To generate complete ground truth masks, we stitched together four navigated cone-beam CT images acquired with the Loop-X immediately after the study and projected organ segmentations~\cite{wasserthal2023totalsegmentator} onto each image.
We observe some spurious masks in the ground truth, possibly due to decomposition, so we limit our evaluation to masks which are at least 2.5\% of the image size. In total, the dataset has 1,741 masks suitable for evaluation.

We find that FluoroSAM effectively segments structures in real X-ray images from a variety of views. With text prompts, FluoroSAM achieves a mean Intersection over Union (IoU) of 0.47 and a Dice score of 0.60, as shown in Table~\ref{tab:results}. This includes hard prompts like ``fifth right rib'' and is sufficient for many downstream applications in interventional imaging~\cite{killeen2025intelligent}. Additionally, although VQ result in marginal improvement for text prompts similar to the training set, we observe qualitative benefits to mask quality for unusual prompts, as seen in Fig.~\ref{fig:ablation}. Fig.~\ref{fig:apps}a shows additional examples with text-only prompting. With 2 points, FluoroSAM outperforms SAM~\cite{kirillov2023segment} (prompted with points or boxes) and MedSAM~\cite{ma2024segment} in terms of IoU and Dice score, achieving 0.56 and 0.69, respectively. In terms of HDD, MedSAM holds a slight advantage over FluoroSAM, possibly because bounding box prompts tightly constrain the mask, whereas point prompts are more ambiguous. As indicated by the lower IoU and Dice, MedSAM generally fails to reflect the underlying anatomy and simply fills the provided box. Moreover, as can be seen in Fig.~\ref{fig:plots}a, point prompting improves the performance of FluoroSAM, whereas SAM predicts increasingly erroneous masks.

\para{Zero-shot evaluation on CXR}
To show FluoroSAM's potential for general diagnostic X-ray image segmentation, we evaluate it on a whole lung segmentation dataset with 1,131 CXRs~\cite{candemir2014lung}. Using only text prompts, FluoroSAM provides a reasonable mask of each lung, achieving a mean IoU of 0.50 and a Dice score of 0.66 (Table~\ref{tab:results}). This is despite training on a dataset of synthetic images with projected masks, which differ systematically from hand-annotated masks~\cite{candemir2014lung}.

\section{Discussion and Conclusion}
\begin{figure}[t]
  \centering
  \includegraphics[width=0.9\linewidth]{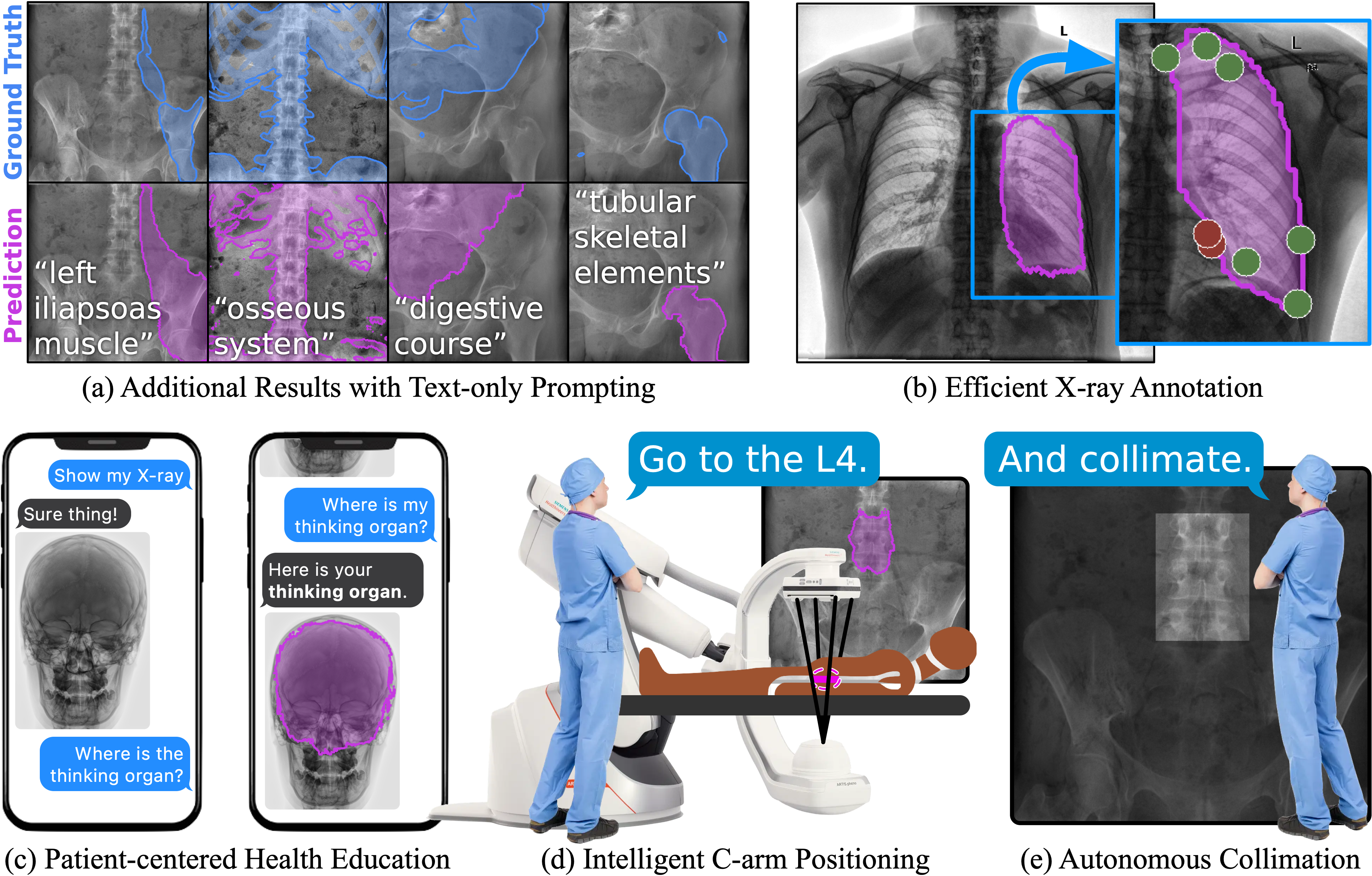}
  \caption{\textbf{Qualitative results and downstream applications.} (a) Additional results on real X-ray images, with text prompts not used during training. The flexibility of text prompting supports a wide variety of downstream applications. (b) Efficient annotation of X-rays can reduce clinical burden and accelerate real data annotation. (c) Flexible text-based prompting may lend itself to patient-facing education, empowering patients to better understand their own anatomy. In the OR, FluoroSAM can be integrated with robotic C-arms to deliver intelligent positioning (d) and autonomous collimation (e), reducing radiation~\cite{killeen2025intelligent}.}
  \label{fig:apps}
\end{figure}

Thanks to their flexibility, LFMs potentially support a wide variety of autonomous and human-in-the-loop downstream applications. As shown in Fig.~\ref{fig:apps}, FluoroSAM's text and point-prompting capabilities lend themselves to multiple areas within a patient's healthcare journey, from efficient annotation of diagnostic images to accessible tele-health tools, equipped to interpret natural language while interacting with exams. In fact, \cite{killeen2025intelligent} demonstrates how to use FluoroSAM to equip robotic C-arms with autonomous positioning and collimation capabilities. A mature FluoroSAM may enable chain-of-thought-based X-ray image analysis related to anatomy in the image~\cite{jiang2024comt}. Here, we have demonstrated FluoroSAM's ability to segment a wide range of anatomical and non-anatomical objects in synthetic, interventional, and chest X-rays. This is a challenging task, as many organs are not plainly visible in X-ray images, which often contain overlapping projections of multiple objects.
Nonetheless, with the ability to interpret a wide range of prompts, we envision FluoroSAM will become a key enabler for human-machine interaction in the X-ray domain, opening up new avenues in both diagnostic and interventional imaging.








\bibliographystyle{splncs04}
\ifthenelse{\equal{\jobname}{\detokenize{final}}}
{
  \bibliography{references-short}
}{
  \bibliography{references}

\begin{thebibliography}{10}
\providecommand{\url}[1]{\texttt{#1}}
\providecommand{\urlprefix}{URL }
\providecommand{\doi}[1]{https://doi.org/#1}

\bibitem{killeen2025intelligent}
Anonymous: Intelligent control of robotic x-ray devices using a language-promptable digital twin. IPCAI Conference (to appear)  (Jun 2025)

\bibitem{brown2020language}
Brown, T.B., Mann, B., Ryder, N., Subbiah, M., Kaplan, J., Dhariwal, P., Neelakantan, A., Shyam, P., Sastry, G., Askell, A., Agarwal, S., Herbert-Voss, A., Krueger, G., Henighan, T., Child, R., Ramesh, A., Ziegler, D.M., Wu, J., Winter, C., Hesse, C., Chen, M., Sigler, E., Litwin, M., Gray, S., Chess, B., Clark, J., Berner, C., McCandlish, S., Radford, A., Sutskever, I., Amodei, D.: {Language Models are Few-Shot Learners}. arXiv  (May 2020). \doi{10.48550/arXiv.2005.14165}

\bibitem{candemir2014lung}
Candemir, S., Jaeger, S., Palaniappan, K., Musco, J.P., Singh, R.K., Xue, Z., Karargyris, A., Antani, S., Thoma, G., McDonald, C.J.: {Lung segmentation in chest radiographs using anatomical atlases with nonrigid registration}. IEEE Trans. Med. Imaging  \textbf{33}(2),  577--590 (Feb 2014). \doi{10.1109/TMI.2013.2290491}

\bibitem{calli2021deep}
{\ifmmode\mbox{\c{C}}\else\c{C}\fi}all{\ifmmode\imath\else\i\fi}, E., Sogancioglu, E., van Ginneken, B., van Leeuwen, K.G., Murphy, K.: {Deep learning for chest X-ray analysis: A survey}. Med. Image Anal.  \textbf{72},  102125 (Aug 2021). \doi{10.1016/j.media.2021.102125}

\bibitem{chen2024chexagent}
Chen, Z., Varma, M., Delbrouck, J.B., Paschali, M., Blankemeier, L., Van~Veen, D., Valanarasu, J.M.J., Youssef, A., Cohen, J.P., Reis, E.P., Tsai, E.B., Johnston, A., Olsen, C., Abraham, T.M., Gatidis, S., Chaudhari, A.S., Langlotz, C.: {CheXagent: Towards a Foundation Model for Chest X-Ray Interpretation}. arXiv  (Jan 2024). \doi{10.48550/arXiv.2401.12208}

\bibitem{edgar2020new}
Edgar, H., Daneshvari~Berry, S., Moes, E., Adolphi, N., Bridges, P., Nolte, K.: New mexico decedent image database. Office of the Medical Investigator, University of New Mexico (2020). \doi{10.25827/5s8c-n515}

\bibitem{fan2023improving}
Fan, L., Krishnan, D., Isola, P., Katabi, D., Tian, Y.: {Improving CLIP Training with Language Rewrites}. arXiv  (May 2023). \doi{10.48550/arXiv.2305.20088}

\bibitem{gao2023synthetic}
Gao, C., Killeen, B.D., Hu, Y., Grupp, R.B., Taylor, R.H., Armand, M., Unberath, M.: {Synthetic data accelerates the development of generalizable learning-based algorithms for X-ray image analysis}. Nat. Mach. Intell.  \textbf{5}(3),  294--308 (Mar 2023). \doi{10.1038/s42256-023-00629-1}

\bibitem{irvin2019chexpert}
Irvin, J., Rajpurkar, P., Ko, M., Yu, Y., Ciurea-Ilcus, S., Chute, C., Marklund, H., Haghgoo, B., Ball, R., Shpanskaya, K., Seekins, J., Mong, D.A., Halabi, S.S., Sandberg, J.K., Jones, R., Larson, D.B., Langlotz, C.P., Patel, B.N., Lungren, M.P., Ng, A.Y.: {CheXpert: A Large Chest Radiograph Dataset with Uncertainty Labels and Expert Comparison}. arXiv  (Jan 2019). \doi{10.48550/arXiv.1901.07031}

\bibitem{jiang2024comt}
Jiang, Y., Chen, J., Yang, D., Li, M., Wang, S., Wu, T., Li, K., Zhang, L.: {CoMT: Chain-of-Medical-Thought Reduces Hallucination in Medical Report Generation}. arXiv  (Jun 2024). \doi{10.48550/arXiv.2406.11451}

\bibitem{killeen2024take}
Killeen, B.D., Chaudhary, S., Osgood, G., Unberath, M.: {Take a shot! Natural language control of intelligent robotic X-ray systems in surgery}. Int. J. CARS  \textbf{19}(6),  1165--1173 (Jun 2024). \doi{10.1007/s11548-024-03120-3}

\bibitem{killeen2023silico}
Killeen, B.D., Cho, S.M., Armand, M., Taylor, R.H., Unberath, M.: {In silico simulation: a key enabling technology for next-generation intelligent surgical systems}. Prog. Biomed. Eng.  \textbf{5}(3),  032001 (May 2023). \doi{10.1088/2516-1091/acd28b}

\bibitem{killeen2023corridors}
Killeen, B.D., Gao, C., Oguine, K.J., Darcy, S., Armand, M., Taylor, R.H., Osgood, G., Unberath, M.: {An autonomous X-ray image acquisition and interpretation system for assisting percutaneous pelvic fracture fixation}. Int. J. CARS pp.~1--8 (May 2023). \doi{10.1007/s11548-023-02941-y}

\bibitem{killeen2023pelphix}
Killeen, B.D., Zhang, H., Mangulabnan, J., Armand, M., Taylor, R.H., Osgood, G., Unberath, M.: {Pelphix: Surgical Phase Recognition from X-Ray Images in Percutaneous Pelvic Fixation}. In: {Medical Image Computing and Computer Assisted Intervention {\textendash} MICCAI 2023}, pp. 133--143. Springer, Cham, Switzerland (Oct 2023). \doi{10.1007/978-3-031-43996-4_13}

\bibitem{kirillov2023segment}
Kirillov, A., Mintun, E., Ravi, N., Mao, H., Rolland, C., Gustafson, L., Xiao, T., Whitehead, S., Berg, A.C., Lo, W.Y., Doll{\ifmmode\acute{a}\else\'{a}\fi}r, P., Girshick, R.: {Segment Anything}. arXiv  (Apr 2023). \doi{10.48550/arXiv.2304.02643}

\bibitem{Liu2023CVPR}
Liu, X., Peng, H., Zheng, N., Yang, Y., Hu, H., Yuan, Y.: Efficientvit: Memory efficient vision transformer with cascaded group attention. In: Proceedings of the IEEE/CVF Conference on Computer Vision and Pattern Recognition (CVPR). pp. 14420--14430 (June 2023)

\bibitem{luijten20233d}
Luijten, G., Gsaxner, C., Li, J., Pepe, A., Ambigapathy, N., Kim, M., Chen, X., Kleesiek, J., H{\ifmmode\ddot{o}\else\"{o}\fi}lzle, F., Puladi, B., Egger, J.: {3D surgical instrument collection for computer vision and extended reality}. Sci. Data  \textbf{10}(796),  1--12 (Nov 2023). \doi{10.1038/s41597-023-02684-0}

\bibitem{ma2024segment}
Ma, J., He, Y., Li, F., Han, L., You, C., Wang, B.: {Segment anything in medical images}. Nat. Commun.  \textbf{15}(654), ~1--9 (Jan 2024). \doi{10.1038/s41467-024-44824-z}

\bibitem{murphy2020general}
Murphy, A., Bell, D., Knipe, H., et~al.: General radiography curriculum. Radiopaedia.org  (2020). \doi{10.53347/rID-48992}, \url{https://doi.org/10.53347/rID-48992}, reference article, Accessed on 22 Feb 2025

\bibitem{oord2017neural}
Oord, A.v.d., Vinyals, O., Kavukcuoglu, K.: {Neural Discrete Representation Learning}. arXiv  (Nov 2017). \doi{10.48550/arXiv.1711.00937}

\bibitem{radford2021learning}
Radford, A., Kim, J.W., Hallacy, C., Ramesh, A., Goh, G., Agarwal, S., Sastry, G., Askell, A., Mishkin, P., Clark, J., Krueger, G., Sutskever, I.: {Learning Transferable Visual Models From Natural Language Supervision}. arXiv  (Feb 2021). \doi{10.48550/arXiv.2103.00020}

\bibitem{ravi2024sam}
Ravi, N., Gabeur, V., Hu, Y.T., Hu, R., Ryali, C., Ma, T., Khedr, H., R{\ifmmode\ddot{a}\else\"{a}\fi}dle, R., Rolland, C., Gustafson, L., Mintun, E., Pan, J., Alwala, K.V., Carion, N., Wu, C.Y., Girshick, R., Doll{\ifmmode\acute{a}\else\'{a}\fi}r, P., Feichtenhofer, C.: {SAM 2: Segment Anything in Images and Videos}. arXiv  (Aug 2024). \doi{10.48550/arXiv.2408.00714}

\bibitem{silva2018automatic}
Silva, G., Oliveira, L., Pithon, M.: {Automatic segmenting teeth in X-ray images: Trends, a novel data set, benchmarking and future perspectives}. Expert Syst. Appl.  \textbf{107},  15--31 (Oct 2018). \doi{10.1016/j.eswa.2018.04.001}

\bibitem{unberath2019enabling}
Unberath, M., Zaech, J.N., Gao, C., Bier, B., Goldmann, F., Lee, S.C., Fotouhi, J., Taylor, R., Armand, M., Navab, N.: {Enabling machine learning in X-ray-based procedures via realistic simulation of image formation}. Int. J. CARS  \textbf{14}(9),  1517--1528 (Sep 2019). \doi{10.1007/s11548-019-02011-2}

\bibitem{unberath2018deepdrr}
Unberath, M., Zaech, J.N., Lee, S.C., Bier, B., Fotouhi, J., Armand, M., Navab, N.: {DeepDRR {\textendash} A Catalyst for Machine Learning in Fluoroscopy-Guided Procedures}. In: {Medical Image Computing and Computer Assisted Intervention {\textendash} MICCAI 2018}, pp. 98--106. Springer, Cham, Switzerland (Sep 2018). \doi{10.1007/978-3-030-00937-3_12}

\bibitem{wasserthal2023totalsegmentator}
Wasserthal, J., Breit, H.C., Meyer, M.T., Pradella, M., Hinck, D., Sauter, A.W., Heye, T., Boll, D.T., Cyriac, J., Yang, S., Bach, M., Segeroth, M.: {TotalSegmentator: Robust Segmentation of 104 Anatomic Structures in CT Images}. Radiology: Artificial Intelligence  (Jul 2023), \url{https://pubs.rsna.org/doi/10.1148/ryai.230024}

\bibitem{zhao2021artificial}
Zhao, W., Shen, L., Islam, M.T., Qin, W., Zhang, Z., Liang, X., Zhang, G., Xu, S., Li, X.: {Artificial intelligence in image-guided radiotherapy: a review of treatment target localization}. Quantitative Imaging in Medicine and Surgery  \textbf{11}(12), ~4881 (Dec 2021). \doi{10.21037/qims-21-199}

\end{thebibliography}
}
\end{document}